\documentclass[journal]{IEEEtran}

\usepackage{cite}
\usepackage{color}
\usepackage{subcaption}
\usepackage{subfig}
\usepackage{epstopdf}
\usepackage{color} 
\usepackage{transparent}
\usepackage{url}
\usepackage{multirow}
\usepackage{array}
\usepackage{amssymb}

\pdfoutput=1

\usepackage[pdftex]{graphicx}
\DeclareGraphicsExtensions{.pdf, .eps}
\usepackage{amsmath}
\DeclareMathOperator{\imresize}{imresize}
\DeclareMathOperator{\rank}{rank}
\DeclareMathOperator*{\argmin}{arg\,min}

\makeatletter
\newcommand{\thickhline}{%
    \noalign {\ifnum 0=`}\fi \hrule height 1pt
    \futurelet \reserved@a \@xhline
}
\newcolumntype{"}{@{\hskip\tabcolsep\vrule width 1pt\hskip\tabcolsep}}

\usepackage{arydshln}
\begin{document}

\title{Face Hallucination using Linear Models of Coupled Sparse Support}

\author{Reuben~A.~Farrugia,~\IEEEmembership{Member,~IEEE,}
        and~Christine~Guillemot,~\IEEEmembership{Fellow,~IEEE}
\thanks{Manuscript received December 18, 2015;}
\thanks{R.A.Farrigoa is with the Department of Communications and Computer Engineering, University of Malta, Malta e-mail: (reuben.farrugia@um.edu.mt).}
\thanks{C. Guillemot is with the Institut National de Recherche
en Informatique et en Automatique, Rennes 35042, France (christine.guillemot@inria.fr).}}

\maketitle

\begin{abstract}

Most face super-resolution methods assume that low-resolution and high-resolution manifolds have similar local geometrical structure, hence learn local models on the low-resolution manifolds (e.g. sparse or locally linear embedding models), which are then applied on the high-resolution manifold.
However, the low-resolution manifold is distorted by the one-to-many relationship between low- and high- resolution patches. 
This paper presents a method which learns linear models based on the local geometrical structure on the high-resolution manifold rather than on the low-resolution manifold.
For this, in a first step, the low-resolution patch is used to derive a globally optimal estimate of the high-resolution patch.
The approximated solution is shown to be close in Euclidean space to the ground-truth but is generally smooth and lacks the texture details needed by state-of-the-art face recognizers.
This first estimate allows us to find the support of the high-resolution manifold using sparse coding (SC), which are then used as support for learning a local projection (or upscaling) model between the low-resolution and the high-resolution manifolds using Multivariate Ridge Regression (MRR).
Experimental results show that the proposed method outperforms six face super-resolution methods in terms of both recognition and quality.
These results also reveal that the recognition and quality are significantly affected by the method used for stitching all super-resolved patches together, where quilting was found to better preserve the texture details which helps to achieve higher recognition rates. 

\end{abstract}

%

\IEEEpeerreviewmaketitle

\section{Introduction}
\label{sec:intro}

\IEEEPARstart{M}{ost} countries around the world use Closed Circuit Television (CCTV) systems to combat crime in their major cities.
These cameras are normally installed to cover a large field of view where the query face image may not be sampled densely enough by the camera sensors \cite{zou_2012}.
The low-resolution and quality of face images captured on camera reduces the effectiveness of CCTV in identifying perpetrators and potential eyewitnesses \cite{sasse_2010, la_vigne_2011}.

Super-resolution techniques can be used to enhance the quality of low-resolution facial images to improve the recognition performance of existing face recognition software and the identification of individuals from CCTV images. 
In a recent survey Wang \textit{et al}. distinguishes between two main categories of  super-resolution methods: reconstruction based and learning based approaches \cite{wang_2014}. 
Reconstruction based methods register a sequence of low-resolution images onto a high-resolution grid and fuse them to suppress the aliasing caused by under-sampling \cite{elad_1999, farsiu_2004}.
On the other hand, learning based methods use coupled dictionaries to learn the mapping relations between low- and high- resolution image pairs to synthesize high-resolution images from low-resolution images \cite{wang_2014,nasrollahi_2014}.
The research community has lately focused on the latter category of super-resolution methods, since they can provide higher quality images and larger magnification factors.

In their seminal work Baker and Kanade \cite{baker_2000} exploited the fact that the human face images are a relatively small subset of natural scenes and introduced the concept of face super-resolution (also known as face hallucination) where only facial images are used to construct the dictionaries. The high-resolution face image is then hallucinated using Bayesian inference with gradient priors. 
The authors in \cite{hu_2011} assume that two similar face images share similar local pixel structures so that each pixel could be generated by a linear combination of spatially neighbouring pixels.
This method was later extended in \cite{li_2014} where they use sparse local pixel structure. 
Although these methods were found to perform well at moderately low-resolutions, they fail when considering very low-resolution face images where the local pixel structure is severely distorted.
Classical face representation models such as Principal Component Analysis (PCA) \cite{wang_2005, park_2008, liu_2001, li_2004, liu_2007}, Kernel PCA (KPCA) \cite{chakrabarti_2007}, Locality Preserving Projections (LPP) \cite{zhuang_2007} and Non-Negative Matrix Factorization (NMF) \cite{yang_2008}, were used to model a novel low-resolution face image using a linear combination of prototype low-resolution face images present in a dictionary. 
The combination weights are then used to combine the corresponding high-resolution prototype face images to hallucinate the high-resolution face image.
Nevertheless, global methods do not manage to recover the local texture details which are essential for face recognition.
To alleviate this problem, some methods employ patch-based local approaches such as Markov Random Fields \cite{liu_2001, li_2004, liu_2007}, Locally Linear Embedding (LLE) \cite{zhuang_2007}, Sparse Coding (SC) \cite{yang_2008} as a post-process.

Different data representation methods, including PCA \cite{chen_2014}, Principal Component Sparse Representation (PCSR)  \cite{lu_2013}, LLE in both spatial \cite{su_2005, liu_2005b} and Discrete Cosine Transform (DCT) domain \cite{zhang_2008, zhang_2011,du_2013}, LPP \cite{park_2007}, Orthogonal LPP \cite{kumar_2008} (OLPP), Tensors \cite{liu_2005}, Constrained Least Squares \cite{ma_2009}, Sparse Representation \cite{jung_2011}, Locality-Constrained Representation \cite{jiang_2012, jiang_2014b}, and Local Appearance Similarity (LAS) \cite{li_2013} have used the same concept to hallucinate high-resolution overlapping patches which are then stitched together to reconstruct the high-resolution face image. 
All these methods \cite{wang_2005, park_2008, liu_2001, li_2004, liu_2007,chakrabarti_2007,zhuang_2007,yang_2008, chen_2014,lu_2013, su_2005, liu_2005b,  zhang_2008, zhang_2011,du_2013,park_2007, kumar_2008, liu_2005,  ma_2009, jung_2011, jiang_2012, jiang_2014b,li_2013 } assume that low- and high- manifolds have similar local geometrical structures.
However, the authors in \cite{su_2005, li_2009a, li_2009b, hao_2014} have shown that this assumption does not hold well because the one-to-many mappings from low- and high-resolution distort the structure of the low-resolution manifold. 

Motivated by this observation, Coupled Manifold Alignment \cite{li_2009b}, Easy-Partial Least Squares (EZ-PLS) \cite{hao_2014}, Dual Associative Learning \cite{liu_2005c}, and  Canonical Correlation Analysis (CCA)\cite{huang_2010},  were used to derive a pair of projection matrices that can be used to project both low- and high-resolution patches on a common coherent subspace.
However, the dimension of the coherent sub-spaces is equal to the lowest rank of the  low- and high-resolution dictionary matrices.
Therefore, the projection from the coherent sub-space to the high-resolution manifold is ill-conditioned.
On the other hand, the Locality-constrained Iterative Neighbour Embedding (LINE) method presented in  \cite{jiang_2013, qu_2014} reduces the dependence from the low-resolution manifold by iteratively updating the neighbours on the high-resolution manifold. 
This was later extended by the same authors in \cite{jiang_2014} where an iterative dictionary learning scheme was integrated to bridge the low-resolution and high-resolution manifolds.
Although this method yields state-of-the-art performance, it cannot guarantee to converge to an optimal solution.
A recent method based on Transport Analysis was proposed in \cite{kolouri_2015} where the high-resolution face image is reconstructed by morphing high resolution training images which best fit the given low-resolution face image.
However, this method heavily relies on the assumption that the degradation function is known, which is generally not possible in typical CCTV scenarios.

Different automated cross-resolution face recognition methods have been proposed to cater for the resolution discrepancy between the gallery and probe images\footnote{Gallery images are high-quality frontal facial images stored in a database which are usually taken in a controlled environment (e.g. ID and passport photos). Probe images are query face images which are compared to each and every face image  included in the gallery. Probe images are usually taken in a non-controlled environment and can have different resolution.}.
These methods either try to include the resolution discrepancy within the classifier's optimization function \cite{yeomans_2008,zou_2012,bhatt_2014,jian_2015} or else by projecting both probe and gallery images on a coherent subspace and compute the classification there \cite{li_2010, zhou_2011, siena_2012, siena_2013}.
However, although these methods are reported to provide good results, they suffer from the following shortcomings \romannumeral 1) most of the methods (\cite{bhatt_2014, li_2010,zhou_2011, siena_2012, siena_2013}) do not synthesize a high resolution face image and \romannumeral 2) they generally assume that multiple images per subject are available in the gallery, which are often scarce in practice.

This work presents a two layer approach named Linear Models of Coupled Sparse Support (LM-CSS),  which employs a coupled dictionary containing low- and corresponding high-resolution training patches to learn the mapping relation between low- and high-resolution patch pairs.
LM-CSS first employs all atoms in the dictionary to project the low-resolution patch onto the high-resolution manifold to get an initial approximation of the high-resolution patch.
This solution gives a good estimate of the high-resolution ground-truth. 
However, experimental results provided in section \ref{sec:background} show that texture detail is better preserved if only a sub-set of coupled low- and high-resolution atoms are used for reconstruction.
Basis Pursuit De-noising (BPDN) was then used to derive the atoms needed to optimally reconstruct the initial approximation on the high-resolution manifold.
Given that high-resolution patches reside on a high-resolution manifold and that the initial solution is sufficiently close to the ground-truth, we
exploit the locality of the high-resolution manifold to refine the initial solution.
This set of coupled sparse support are then used by a Multivariate Ridge Regression (MRR) to model the up-scaling function for each patch, which better preserves the texture detail crucial for recognition.

The proposed approach was extensively evaluated against six face hallucination methods using 930 probe images from the FRGC dataset against a gallery of 889 individuals using a closed set identification scenario with one face image per subject in the gallery\footnote{Collection of gallery images is laborious and expensive in practice. This limits the number of gallery images that can be used in practice for recognition, where frequently only one image per subject is available in the gallery. This problem is referred to as the one sample per person in face recognition literature \cite{tan_2006}.}.
This method was found to provide state-of-the-art performance in terms of face recognition achieving rank-1 recognition gains between 0.4\% - 1.5\% over LINE \cite{jiang_2014} and between 1\% - 9\% over Sparse Position Patching \cite{jung_2011} who ranked second and third respectively.
The quality analysis further shows that the proposed method outperforms Eigen-Patches \cite{chen_2014} and Position-Patches \cite{ma_2009} by around 0.1 dB and 0.2 dB in Peak Signal-to-Noise Ratio (PSNR) respectively, followed by the others.

The remained of the paper is organized as follows
In section \ref{sec:background} we give some notation and concepts needed later in the paper.
The proposed method is described in section \ref{sec:scsRR} while the experimental protocol is outlined in the following section.
The experimental result and discussion are presented in section \ref{sec:results}, while the final comments and conclusion are provided in section \ref{sec:conclusion}.

\section{Background}
\label{sec:background}

\subsection{Problem Formulation}
\label{sec:problem_form}
We consider a low-resolution face image $\mathbf{X}$ where the distance between the eye centres is defined as $d_x$.
The goal of face super-resolution is to up-scale $\mathbf{X}$ by a scale factor $\alpha = \frac{d_y}{d_x}$, where $d_y$ represents the distance between the eye-centres of the desired super-resolved face image.
The image $\mathbf{X}$ is divided into a set of overlapping patches of size $\sqrt{n} \times \sqrt{n}$ with an overlap of $\gamma_x$, and the resulting patches are reshaped to column-vectors in lexicological order and stored as vectors $\mathbf{x}_i$, where $i \in [1,p]$ represents the patch index.

In order to learn the mapping relation between low- and high-resolution patches, we have $m$ high resolution face images which are registered based on eye and mouth center coordinates, where the distance between eye centres is set to $d_y$.
These images are divided into overlapping patches of size $\left[ \alpha \sqrt{n} \times \alpha \sqrt{n} \right]$ with an overlap of $\gamma_y = \left[\alpha \gamma_x \right]$, where $\left[ * \right]$ stands for the rounding operator.
The $i$-th patch of every high-resolution image is reshaped to column-vectors in lexicological order and placed within the high-resolution dictionary $\mathbf{H}_i$.
The low-resolution dictionary of the $i$-th patch $\mathbf{L}_i$ is constructed using the same images present in the high-resolution dictionary, which are down-scaled by a factor $\frac{1}{\alpha}$ and divided into overlapping patches of size $\sqrt{n} \times \sqrt{n}$ with an overlap of $\gamma_x$.
This formulation is in line with the position-patch method published in \cite{ma_2009} where only collocated patches with index $i$ are used to super-resolve the low resolution patch $\mathbf{x}_i$.

Without loss of generalization we will assume that the column vectors of both dictionaries are standardized to have zero mean and unit variance to compensate for illumination and contrast variations.
The standardized low-resolution patch is denoted by $\mathbf{x}_i^s$ and the aim of this work is to find an up-scaling projection matrix that minimizes the following objective function

\begin{equation}
\boldsymbol{\Phi}_i = \argmin_{\boldsymbol{\Phi}_i}{|| \mathbf{H}_i - \boldsymbol{\Phi}_i \mathbf{L}_i||^2_2}
\label{eq:phi_ls_formulation}
\end{equation}

\noindent where $\boldsymbol{\Phi}_i$ is the up-scaling projection matrix of dimensions $[\alpha]^2 n \times n $. The standardized $i$-th high-resolution patch is then hallucinated using

\begin{equation}
\widetilde{\mathbf{y}}_i^s = \boldsymbol{\Phi}_i \mathbf{x}_i^s
\label{eq:hallucination_standardizes}
\end{equation}

The pixel intensities of the patch are then recovered using

\begin{equation}
\widetilde{\mathbf{y}}_i = [ \sigma_i \widetilde{\mathbf{y}}_i^s + \mu_i ]
\label{eq:hallucination}
\end{equation}

\noindent where $\mu_i$ and $\sigma_i$ represent the mean and standard deviation of the low-resolution patch $\mathbf{x}_i$.
The resulting hallucinated patches are then stitched together to form the hallucinated high-resolution face image $\widetilde{\mathbf{Y}}$.

\subsection{Multivariate Ridge Regression}
\label{sec:mvrr}

The direct least squares solution to \eqref{eq:phi_ls_formulation} can provide singular values. Instead, this work adopts an $l_2-$norm regularization term to derive a more stable solution for the up-scaling projection matrix $\boldsymbol{\Phi}_i$.
The analytic solution of multivariate ridge regression is given by

\begin{equation}
\begin{aligned}
\boldsymbol{\Phi}_i &= \argmin_{\boldsymbol{\Phi}_i} {|| \mathbf{H}_i - \boldsymbol{\Phi}_i \mathbf{L}_i||^2_2} + \lambda_r || \boldsymbol{\Phi}_i||_2^2  \\
	&= \mathbf{H}_i \mathbf{L}_i^T \left(\mathbf{L}_i \mathbf{L}_i^T + \lambda_r \mathbf{I}\right)^{-1} 
\label{eq:upscale_solution}
\end{aligned}
\end{equation}

\noindent where $\lambda_r$ is a regularization parameter used to avoid singular values and $\mathbf{I}$ is the identity matrix. This solution will be referred to as the direct up-scaling matrix.

Very recently, the authors in \cite{zhang_2015} have adopted ridge regression to solve a different optimization problem for generic super-resolution, which using our notation can be formulated as

\begin{equation}
\boldsymbol{\beta}_i = \argmin_{\boldsymbol{\beta}_i}{|| \mathbf{x}_i^s - \mathbf{L}_i \boldsymbol{\beta}_i||^2_2 + \lambda || \boldsymbol{\beta}_i||_2^2 }
\label{eq:zhang_formulation}
\end{equation}

\noindent where their intent is to find the reconstruction weight vector $\boldsymbol{\beta}_i$ that is optimal to represent the low-resolution patch $\mathbf{x}_i$ using a weighted combination of the low resolution dictionary $\mathbf{L}_i$.
The same reconstruction weights are then used to combine the high-resolution patches in the coupled dictionary to hallucinate the high-resolution patch using

\begin{equation}
  \begin{aligned}
        \widetilde{\mathbf{y}}_i^\zeta &= \sigma_i \mathbf{H}_i \boldsymbol{\beta}_i + \mu_i\\
                           &= \sigma_i \underbrace{\mathbf{H}_i \left( \mathbf{L}_i^T \mathbf{L}_i + \lambda \mathbf{I}\right)^{-1}  \mathbf{L}_i^T}_{\mathbf{\Phi}^\zeta_i} \mathbf{x}_i^s + \mu_i
  \end{aligned}
  \label{eq:zhang_solution}
\end{equation}

\noindent where $\widetilde{\mathbf{y}}_i^\zeta$ represents the hallucinated solution using the method in \cite{zhang_2015}.
It can be seen that the solution in \eqref{eq:zhang_solution} can be interpreted as up-scaling the low-resolution patch $\mathbf{x}_i^s$ using the up-scaling projection matrix $\mathbf{\Phi}^\zeta_i$. 
This is referred to as the indirect up-scaling matrix.

It can be shown in Appendix  \ref{sec:proof_1} that $\mathbf{\Phi}^\zeta_i = \mathbf{\Phi}_i$.
This leads to the following equivalent solutions to the standardized high-resolution patch

\begin{equation}
\widetilde{\mathbf{y}}_i^s = \mathbf{\Phi}_i \mathbf{x}_i^s = \mathbf{H}_i \boldsymbol{\beta}_i
\label{eq:phi_beta_equiv}
\end{equation}

This shows that finding the optimal reconstruction weights to model $\mathbf{x}_i$ and then use them to reconstruct the high-resolution patch (indirect method) is equivalent to modelling the up-scaling projection matrix directly (as the proposed method). 
Nevertheless, the direct formulation has a complexity of the order $O((1+\alpha^2)n^2(n+m)) \approx O(n^3)$ while the indirect formulation has a complexity of the order $O(\alpha^2n^2m + m^3 + m^2n) \approx O(m^3)$, where $m \gg n$.

\section{Linear Models of Coupled Sparse Support}
\label{sec:scsRR}

\subsection{Quality and Texture Analysis}
\label{sec:quality_texture_analysis}

\begin{figure*}[htb]
\begin{subfigure}{.5\textwidth}
  \centering
  \includegraphics[width=.95\linewidth]{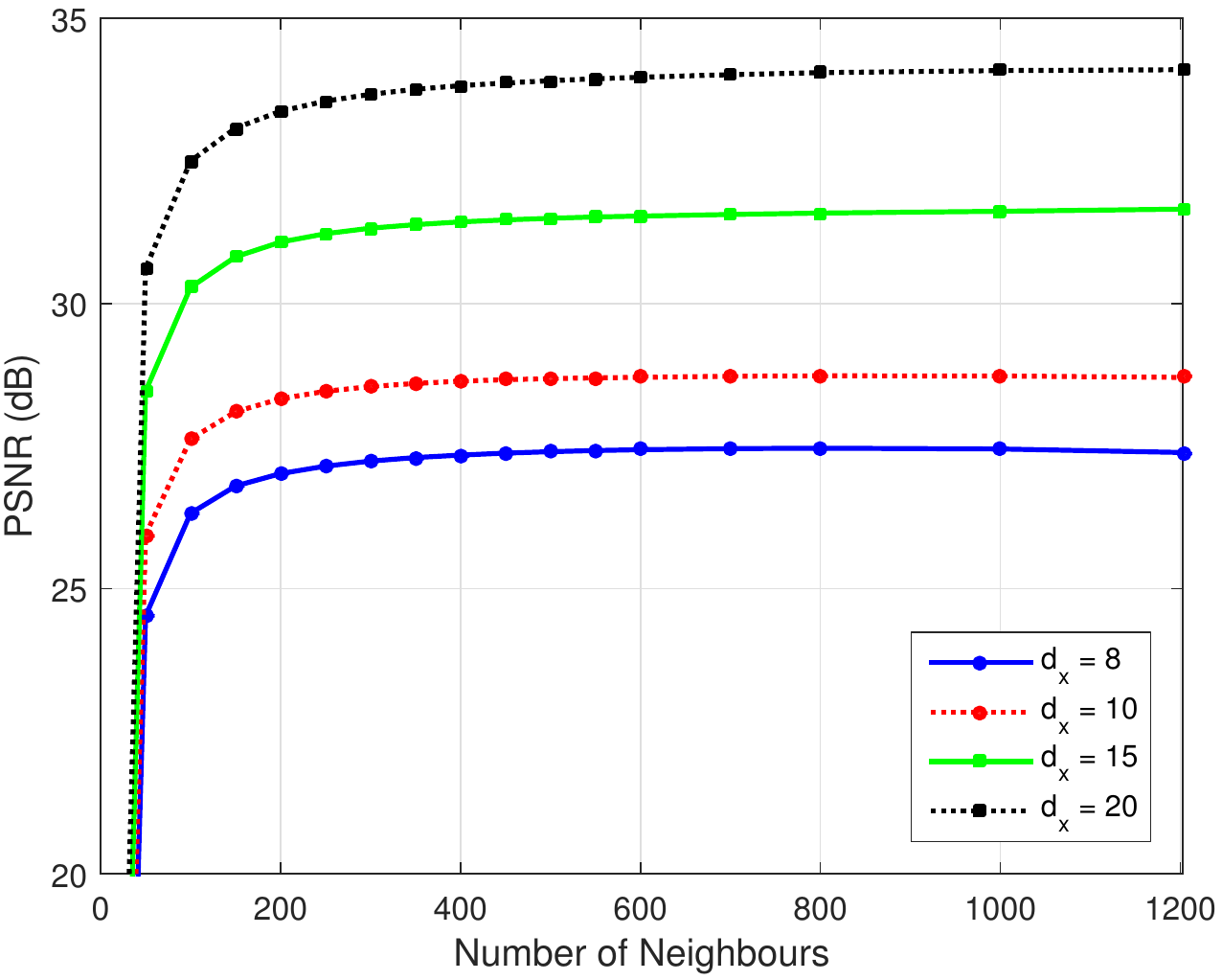}
  \caption{Quality Analysis}
  \label{fig:quality_analysis}
\end{subfigure}%
\begin{subfigure}{.5\textwidth}
  \centering
  \includegraphics[width=.95\linewidth]{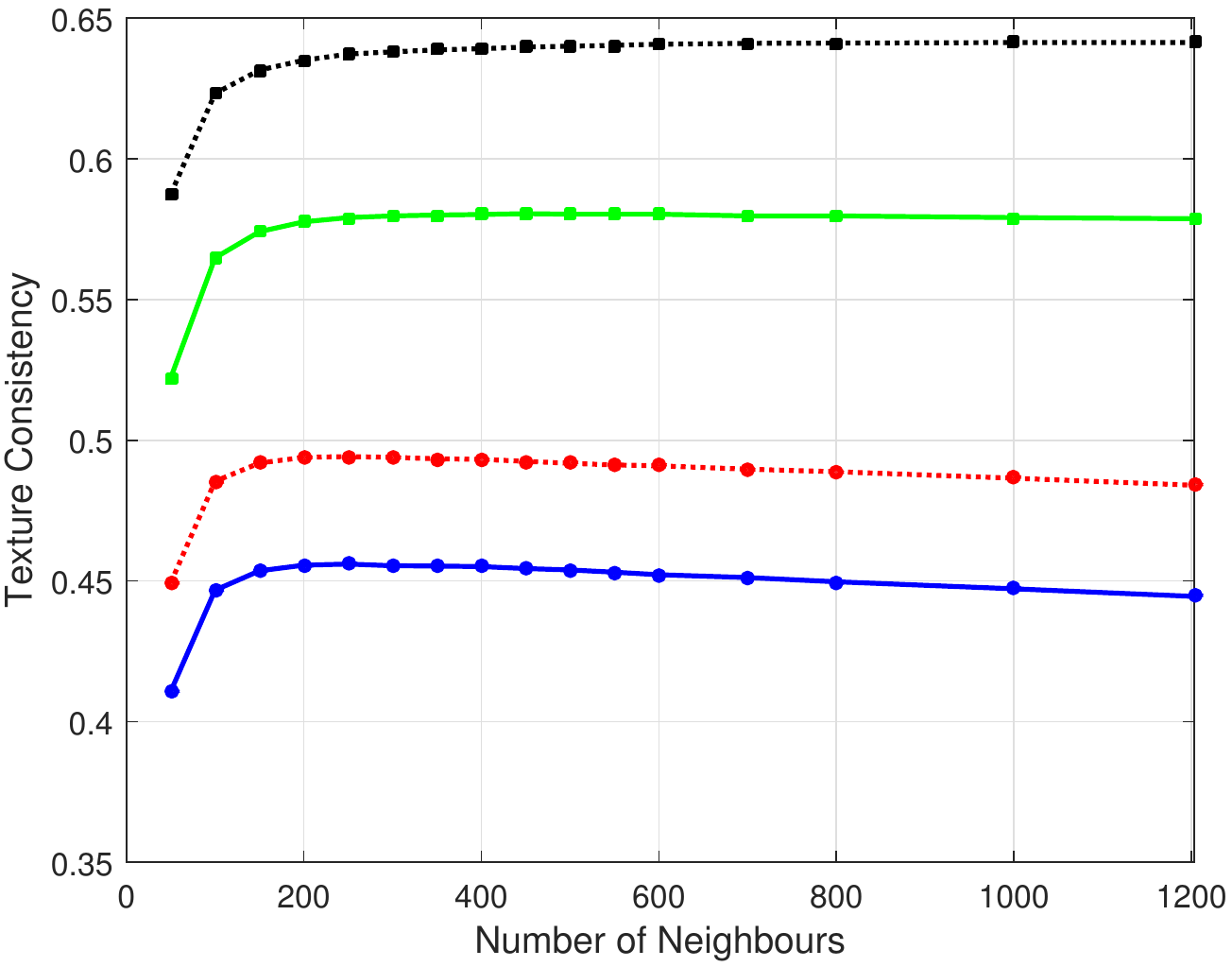}
  \caption{Texture Analysis}
  \label{fig:texture_analysis}
\end{subfigure}
\caption{Performance analysis using different neighbourhood size $k$.}
\label{fig:quality_texture_analysis}
\end{figure*}

The multivariate ridge regression solution given in \eqref{eq:upscale_solution} is heavily over-determined since it employs all $m$ column-vectors.
This generally results in biased and overly-smooth solutions \cite{jung_2011}.
Motivated by the success of Neighbour Embedding based schemes, we will investigate here the effect that neighbourhood size has on performance.
We define $\mathbf{s}$ to represent the support which specifies the column-vector indices that are used to model the up-scaling projection matrix.
Let $\mathbf{L}_i(\mathbf{s})$ and $\mathbf{H}_i(\mathbf{s})$ represent the coupled sub-dictionaries using the column-vectors listed in $\mathbf{s}$ from $\mathbf{L}_i$ and $\mathbf{H}_i$ respectively.
Therefore, the up-scaling projection matrix can be solved using

\begin{equation}
  \begin{aligned}
  	\mathbf{\Phi}_i^k &= \argmin_{\mathbf{\Phi}_i^k} { || \mathbf{H}_i(\mathbf{s}) - \mathbf{\Phi}_i^k} \mathbf{L}_i(\mathbf{s})||_2^2 + \lambda || \mathbf{\Phi}_i^k||_2^2 \\
  	&= \mathbf{H}_i(\mathbf{s}) \mathbf{L}_i(\mathbf{s})^T \left( \mathbf{L}_i(\mathbf{s}) \mathbf{L}_i(\mathbf{s})^T + \lambda \mathbf{I}\right)^{-1}
  	  \end{aligned}
\label{eq:upscale_solution_sparse}
\end{equation}

\noindent where $k = |\mathbf{s}|$ represents the cardinality of set $\mathbf{s}$ which corresponds to the size of the support. Figure \ref{fig:quality_texture_analysis} shows the quality and texture analysis as a function of the number of support points $k$.
These results were computed using the AR face dataset \cite{martinez_1998} while the coupled dictionary is constructed using one-image per subject from the Color Feret \cite{philips_1998} and Multi-Pie \cite{gross_2010} datasets.
More information about these datasets is provided in section \ref{sec:experimental_protocol}.
\noindent

In this experiment the support was chosen using $k-$Nearest neighbours. 
The quality was measured using the PSNR\footnote{Similar results were obtained using other full-reference quality metrics such as Structural Similarity (SSIM) and Feature Similarity (FSIM) metrics.} while the Texture Consistency (TC) was measured by comparing the Local Binary Pattern (LBP) features of the reference and hallucinated image. 
The LBP features were extracted using the method in \cite{ahonen_2006} where the similarity was measured using histogram intersection.
In this experiment $n = 25$, $\gamma_x = 2$ and $m = 1203$. More information is provided in section \ref{sec:experimental_protocol}.

\begin{figure*}[htb]
\begin{center}
\begin{tabular}{@{}cccc@{}}
\includegraphics[height=30mm]{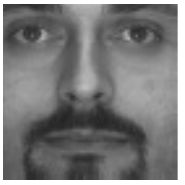} &
\includegraphics[height=30mm]{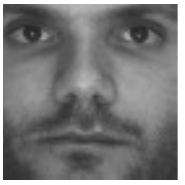} & 
\includegraphics[height=30mm]{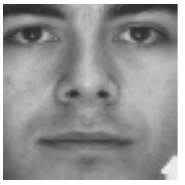} & 
\includegraphics[height=30mm]{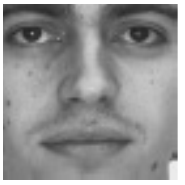} \\ 
\includegraphics[height=30mm]{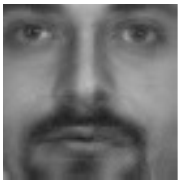} &
\includegraphics[height=30mm]{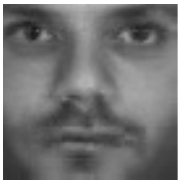} & 
\includegraphics[height=30mm]{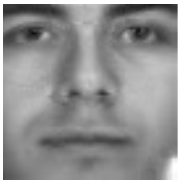} & 
\includegraphics[height=30mm]{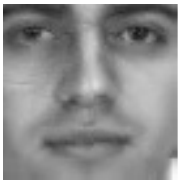} \\ 
\small{PSNR = 29.48 TS = 0.52} & \small{PSNR = 30.23 TS = 0.50} & \small{PSNR = 28.95 TS = 0.55} & \small{PSNR = 29.25 TS = 0.54}\\
\includegraphics[height=30mm]{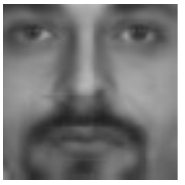} &
\includegraphics[height=30mm]{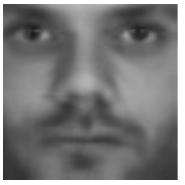} & 
\includegraphics[height=30mm]{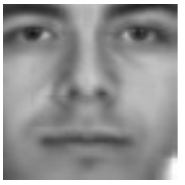} & 
\includegraphics[height=30mm]{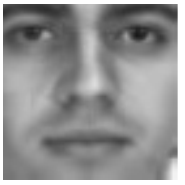} \\ 
\small{PSNR = 30.21 TS = 0.49} & \small{PSNR = 31.03 TS = 0.48} & \small{PSNR = 29.08 TS = 0.53} & \small{PSNR = 29.58 TS = 0.51}\\
\end{tabular}
\end{center}
\caption{Comparison of super-resolution results at resolution $d_x = 10$ where the top row represents the high-resolution ground truth, middle row the hallucination with $k = 150$ and bottom row the hallucination with $k = 1203$.}
\label{fig:subjective_analysis}
\end{figure*}

The results in Figure \ref{fig:quality_analysis} demonstrate that the PSNR increases rapidly until $k = 200$, and keeps on improving slowly for larger values of $k$. 
The up-scaling function $\mathbf{\Phi}_i^k$ which maximizes the PSNR metric, and is therefore closer to the ground-truth in Euclidean space, was obtained when $k = m$ i.e. all column-vectors are included as support.
However, the results in Figure \ref{fig:texture_analysis} show that the texture consistency increases steadily up till $k = 200$ and starts degrading (or remains steady) as $k$ increases.
The subjective results in Figure \ref{fig:subjective_analysis} support this observation where it can be seen that the images derived using $k = 150$ (middle row) generally contain more texture detail while the  images for $k = 1203$ (bottom row) are more blurred.

All the face hallucination methods found in literature follow the same philosophy of generic super-resolution and are designed to maximize an objective measure such as PSNR.
These methods assume that increasing the PSNR metric will inherently improve the face recognition performance.
The above results and observations reveal that improving the PSNR does not correspond to improving the texture detail of the hallucinated face image.
However, recent face recognition methods exploit the texture similarity between probe and gallery images to achieve state-of-the-art performance \cite{ahonen_2006, wagner_2012}.
This indicates that optimizing the face hallucination to minimize the mean square error leads to sub-optimal solutions, at least in terms of recognition.

The results in Fig. \ref{fig:texture_analysis} further show that there is a relation between texture consistency and sparsity, i.e. facial images hallucinated using the $k$-nearest atoms, where $(k \ll m)$, have better texture consistency.
The aim of this work is to learn upscaling models between the low- and high-resolution manifolds exploiting the local geometrical structure of the high-resolution manifold.
Opposed to method proposed by Jung \cite{jung_2011}, Sparse Coding is employed on the high-resolution manifold and the up-scaling matrix is learned using Multivariate Ridge Regression.
The results in section \ref{sec:results} reveal the superiority of the proposed approach over this method.

\subsection{Proposed Method}
\label{sec:proposed_method}

\begin{figure*}[tb] 
\centering 
\def\svgwidth{450pt} 
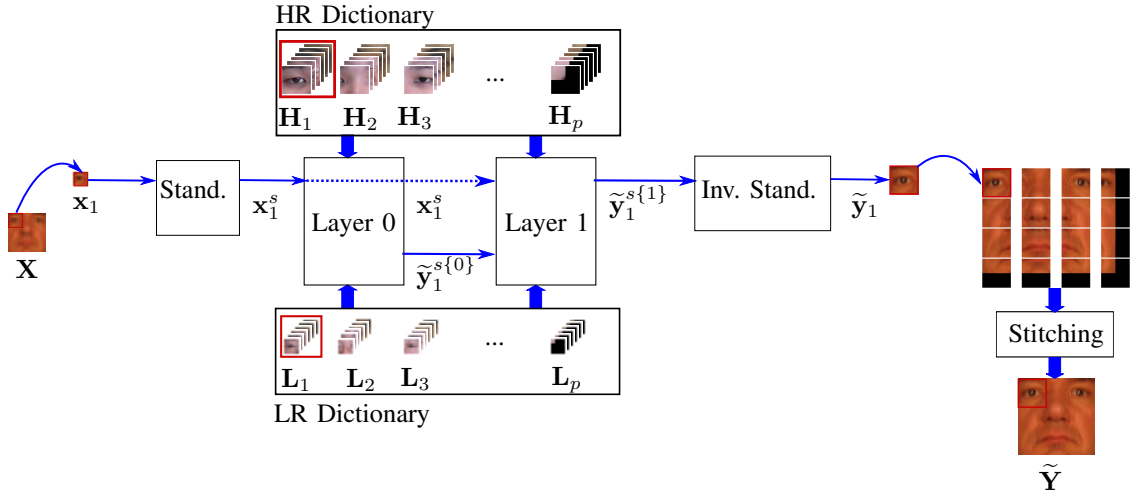 
\caption{Block Diagram of the proposed method.}
\label{fig:overview}
\end{figure*}

The proposed method builds on the observations drawn in the previous sub-section where the main objective is to find the atoms that are able to better preserve the similarity between the hallucinated and ground-truth images in terms of both texture and quality.
Fig. \ref{fig:overview} shows a block-diagram of the proposed method, where in this example the first patch covering the right eye is being processed.
The low-resolution patch $\mathbf{x}_i$ is first standardized and then passed to the first layer which derives the first approximation ${\widetilde{\mathbf{y}}}_i^{s\{0\}}$ of the desired standardized ground-truth ${\mathbf{y}}_i^s$, which is not known in practice.

\noindent

\begin{figure}[tb] 
\centering 
\def\svgwidth{250pt} 
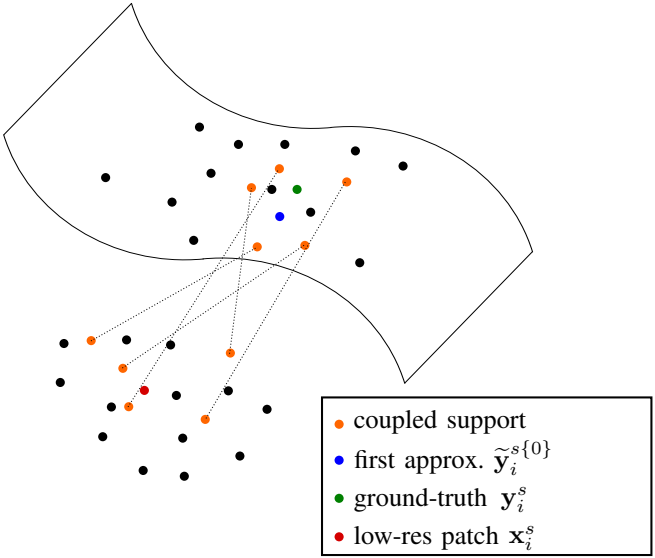 
\caption{Illustration of the concept behind this work.}
\label{fig:general_concept}
\end{figure}

Fig. \ref{fig:general_concept} depicts the geometrical representation of this method and shows that if ${\widetilde{\mathbf{y}}}_i^{s\{0\}}$ is sufficiently close to the ground-truth, they will share similar local-structure on the high-resolution manifold \cite{wang_2010}.
The first approximation is computed using

\begin{equation}
\begin{aligned}
{\widetilde{\mathbf{y}}}_i^{s\{0\}} &= \mathbf{\Phi}_i \mathbf{x}_i^s \\
&= \mathbf{H}_i \mathbf{L}_i^T \left(\mathbf{L}_i \mathbf{L}_i^T + \lambda \mathbf{I}\right)^{-1} \mathbf{x}_i^s
\end{aligned}
\label{eq:initial_approximation}
\end{equation}
\noindent where $\mathbf{\Phi}_i$ is approximated using all atoms in the coupled dictionaries.
This solution ignores the local geometric structure of the low-resolution manifold which is known to be distorted, and tries to approximate the ground-truth vector using the global structure of the low-resolution manifold.
This provides a unique and global solution for the approximation of the ground-truth point.
Backed up by the results in Fig. \ref{fig:quality_analysis}, this solution provides the largest PSNR and is thus close to the ground-truth in Euclidean space. 
However, as shown in Fig. \ref{fig:subjective_analysis}, the solution is generally blurred and lacks important texture details.

The second layer assumes that ${\widetilde{\mathbf{y}}}_i^{s\{0\}}$ is sufficiently close to the ground-truth and exploits the locality of the high-resolution manifold to refine the first approximated solution and recover the required texture details.
The aim here is to find the atoms from the high-resolution dictionary which can optimally reconstruct the first approximation ${\widetilde{\mathbf{y}}}_i^{s\{0\}}$.
This can be formulated as finding the sparsest representation of ${\widetilde{\mathbf{y}}}_i^s$ using 

\begin{equation}
\boldsymbol{\eta}_i = \argmin_{\boldsymbol{\eta}_i}{||\boldsymbol{\eta}_i||_0} \text{ subject to } ||{\widetilde{\mathbf{y}}}_i^s - \mathbf{H}_i \boldsymbol{\eta}_i||_2^2 \leq \delta
\label{eq:sparse_eqn_l0}
\end{equation}

\noindent where $\boldsymbol{\eta}_i$ is the sparse vector, $||\boldsymbol{\eta}_i||_0$ represents the number of non-zero entries in  $\boldsymbol{\eta}_i$ and $\delta$ is the noise parameter.
This solution however is intractable since it cannot be solved in polynomial time.
The authors in \cite{donoho_2001, donoho_2006} have shown that \eqref{eq:sparse_eqn_l0} can be relaxed and solved using Basis Pursuit Denoising (BPDN)

\begin{equation}
\boldsymbol{\eta}_i = \argmin_{\boldsymbol{\eta}_i}{||{\widetilde{\mathbf{y}}}_i^s - \mathbf{H} \boldsymbol{\eta}_i||_2^2 + \lambda_s ||\boldsymbol{\eta}_i||_1} 
\label{eq:sparse_eqn_l1}
\end{equation}

\noindent where $\lambda_s$ is a regularization parameter.
This optimization can be solved in polynomial time using linear programming.
In this work we use the solver provided by SparseLab\footnote{The code can be found at \url{https://sparselab.stanford.edu/}} to solve the above BPDN problem.
The support $\mathbf{s}$ is then set as the  $k$ indices of $\boldsymbol{\eta}_i$ with the largest magnitude.
The high-resolution patch is then solved using

\begin{equation}
\widetilde{\mathbf{y}}_i^s = \widetilde{\mathbf{y}}_i^{s\{1\}} = \mathbf{H}_i(\mathbf{s}) \mathbf{L}_i(\mathbf{s})^T \left( \mathbf{L}_i(\mathbf{s}) \mathbf{L}_i(\mathbf{s})^T + \lambda_r \mathbf{I}\right)^{-1}\mathbf{x}_i^s
\end{equation}
It is important to notice that using $\boldsymbol{\eta}_i$ directly will provide a solution close to the first approximation and therefore the coefficients $\boldsymbol{\eta}_i$ cannot be used directly.
Instead we use the coupled anchor points to get an up-scaling projection matrix with richer texture.
This can be further explained that by extending \eqref{eq:phi_beta_equiv} for the support $\mathbf{s}$, one gets the following relation

\begin{equation}
\mathbf{\Phi}_i^k \mathbf{x}_i^s = \mathbf{H}_i(\mathbf{s}) \boldsymbol{\beta}_i^k
\label{eq:phi_beta_equiv_2}
\end{equation}

\noindent where

\begin{equation}
\boldsymbol{\beta}_i^k = \argmin_{\boldsymbol{\beta}_i^k}{|| \mathbf{x}_i^s - \mathbf{L}_i(\mathbf{s}) \boldsymbol{\beta}_i^k||^2_2} + \lambda ||\boldsymbol{\beta}_i^k ||^2_2
\end{equation}

These equations indicate that a larger support will lead to averaging a larger number of high-resolution atoms, which will result in blurred solutions.
Therefore, increasing the support will reduce the ability of the projection matrix $\mathbf{\Phi}_i^k$ to recover texture details.

\begin{figure*}[htb]
\begin{subfigure}{.5\textwidth}
  \centering
  \includegraphics[width=.95\linewidth]{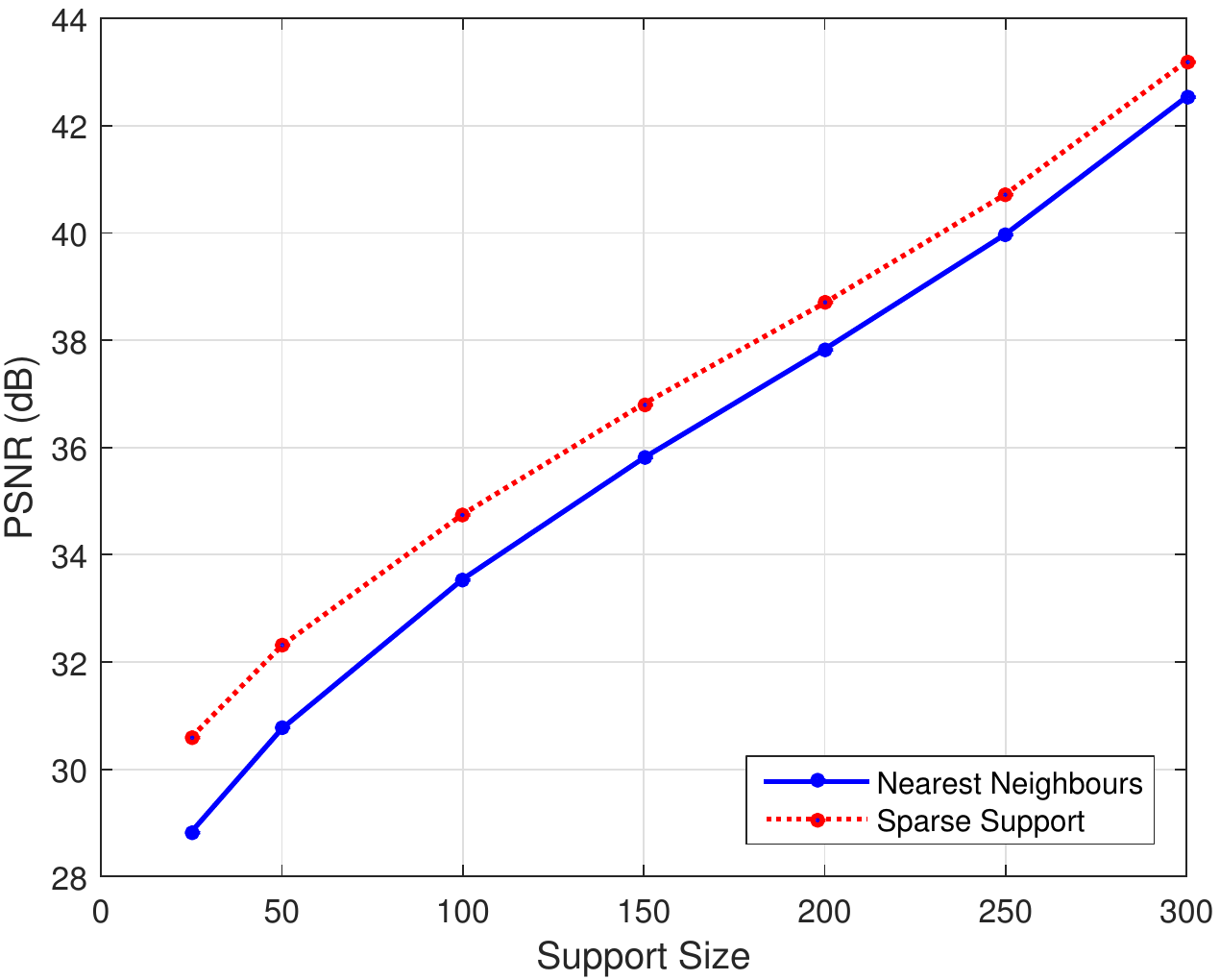}
  \caption{Quality Analysis}
  \label{fig:sparse_quality_analysis}
\end{subfigure}%
\begin{subfigure}{.5\textwidth}
  \centering
  \includegraphics[width=.95\linewidth]{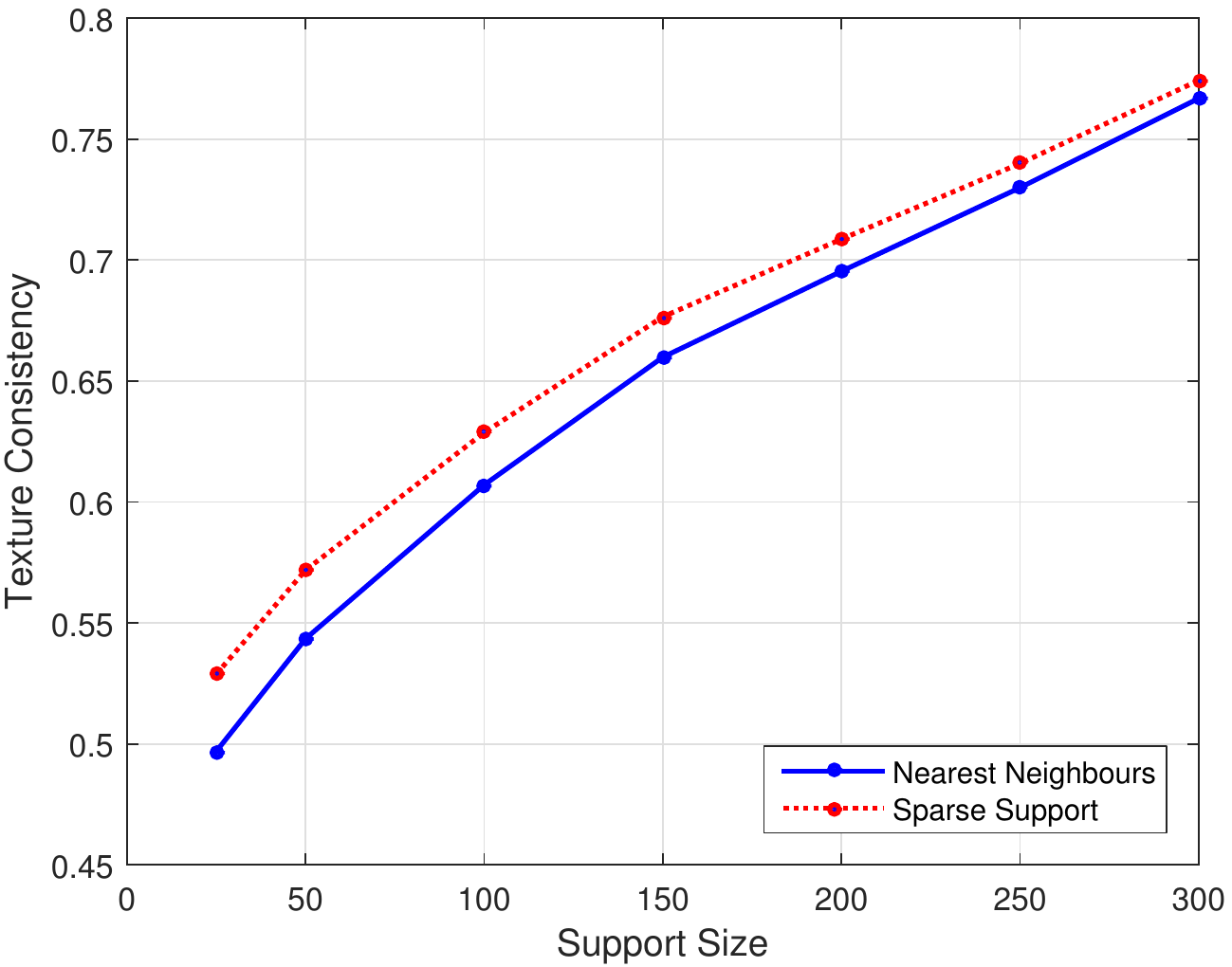}
  \caption{Texture Analysis}
  \label{fig:sparse_texture_analysis}
\end{subfigure}
\caption{Comparing the modelling ability using two different support selection to model the ground-truth samples when the high-resolution patch size is set to 20.}
\label{fig:sparse_quality_texture_analysis}
\end{figure*}

Fig. \ref{fig:sparse_quality_texture_analysis} shows the performance of the proposed sparse support selection method compared to $k$-nearest neighbour, to model the ground-truth samples $\mathbf{y}_i$ using a weighted combination of selected atoms from dictionary $\mathbf{H}_i$. The weights were derived using

\begin{equation}
\boldsymbol{\nu}_i = \left( {\mathbf{H}_i(\mathbf{s})}^T \mathbf{H}_i(\mathbf{s}) \right)^{\dagger}{\mathbf{H}_i(\mathbf{s})}^T \mathbf{y}_i
\end{equation}

\noindent where $\dagger$ is the pseudo-inverse operator.
These results were tested on the AR dataset using similar configuration provided above.
It can be seen that the reconstruction of the ground-truth $\mathbf{y}_i$ using $k$ sparse support  provides images of better quality and texture-consistency than when using the $k$ nearest neighbours.
Therefore, the proposed method finds an optimal initial guess in $l_2$ sense and then finds an optimal set of support using Basis Pursuit Denoising. 

\noindent

The last step involves stitching the overlapping patches together. 
This process is generally given little importance in literature, where most  methods simply average the overlapping regions.
The quilting method \cite{efros_2001} is an alternative approach which tries to find the minimum error boundary patch and better preserves the texture recovered by each patch when reconstructing the global face.
The results in section \ref{sec:results} show that the performance of face hallucination is significantly dependent on the stitching method used.

\section{Experimental Protocol}
\label{sec:experimental_protocol}

The experiments conducted in this paper use different publicly available datasets: \romannumeral 1) AR  \cite{martinez_1998}, \romannumeral 2) Color Feret \cite{philips_1998}, \romannumeral 3) Multi-Pie \cite{gross_2010}, \romannumeral 4) MEDS-II \cite{founds_2011} and \romannumeral 5) FRGC-V2 \cite{phillips_2005}. The AR dataset was primarily used for the analysis provided in section \ref{sec:scsRR}, and consists of 112 images with 8 images per subject resulting in a total of 886 images with different expressions.

The dictionary used to learn the up-scaling projection matrix for each patch consisted of a composite dataset which includes images from both Color Feret and Multi-Pie datasets, where only frontal facial images were considered. One image per subject was randomly selected, resulting in a dictionary of $m =1203$ of facial images. 
The gallery consisted of another composite dataset which combined frontal facial images from the FRGC-V2 (controlled environment) and MEDS datasets.
One unique image per subject was randomly selected, providing a gallery of 889 facial images.
The probe images were taken from the FRGC-V2 dataset (uncontrolled environment), where two images per subjects were included, resulting in 930 probe images.
All probe images are frontal images, however various poses and illuminations were considered.
The performance of the proposed method is reported on a closed set identification scenario.

All the images were registered using affine transformation computed on landmark points of the eyes and mouth centres, such that the distance between the eyes $d_y = 40$.
The probe and low-resolution dictionary images were down-sampled to the desired scale $\alpha$ using MATLAB's $\imresize$ function. 
Unless stated otherwise all patch based methods are configured such that the number of pixels in a low-resolution patch $n = 25$, the low-resolution overlap $\gamma_x = 2$, and the patches are stitched by averaging overlapping regions.
The dictionaries for each patch are constructed using position-patch method introduced in \cite{ma_2009} and described in section \ref{sec:problem_form}.

Two face recognition methods were adopted in this experiment, namely the LBP face recognition \cite{ahonen_2006} method (which was found to provide state-of-the-art performance on the single image per subject problem in \cite{wagner_2012}) and the Gabor face recognizer \cite{struc_2010}\footnote{The code was provided by the authors in \url{http://www.mathworks.com/matlabcentral/fileexchange/35106-the-phd-face-recognition-toolbox}.}.
The latter method performs classification on the Principal Component Analysis (PCA) subspace.
The PCA basis were trained off-line on the AR dataset.
The proposed method was compared with Bi-Cubic Interpolation, a global face method \cite{wang_2005} and five patch-based methods which are reputed to provide state-of-the-art performance \cite{chen_2014, ma_2009, jung_2011, chang_2004, jiang_2014}. 
These methods were configured using the same patch size and overlap as indicated above and configured using the optimal parameters provided in their respective papers.
The methods were implemented in MATLAB, where the code for \cite{jiang_2014} was provided by the authors.
All simulations were run using a machine with Intel (R) Core (TM) i7-3687U CPU at 2.10GHz running Windows 64-bit Operating system.

The proposed method has only three parameters that need to be tuned. The regularization parameter $\lambda_r$ adopted by Ridge Regression can be easily set to a very small value since its purpose is to perturb the linear-dependent vectors within a matrix to avoid singular values.
In all experiments, this was set to $10^{-6}$.
Similarly, the BPDN's regularization parameter $\lambda_s$ which controls the sparsity of the solution was set to 0.01, since it provided satisfactory performance on the AR dataset.
The performance of the proposed method is mainly affected by the number of anchor vectors $k$, where its effect will be extensively analysed in the following section.

\section{Results}
\label{sec:results}

\subsection{Parameter Analysis}

\begin{figure*}[htb]
\begin{subfigure}{.5\textwidth}
  \centering
  \includegraphics[width=.95\linewidth]{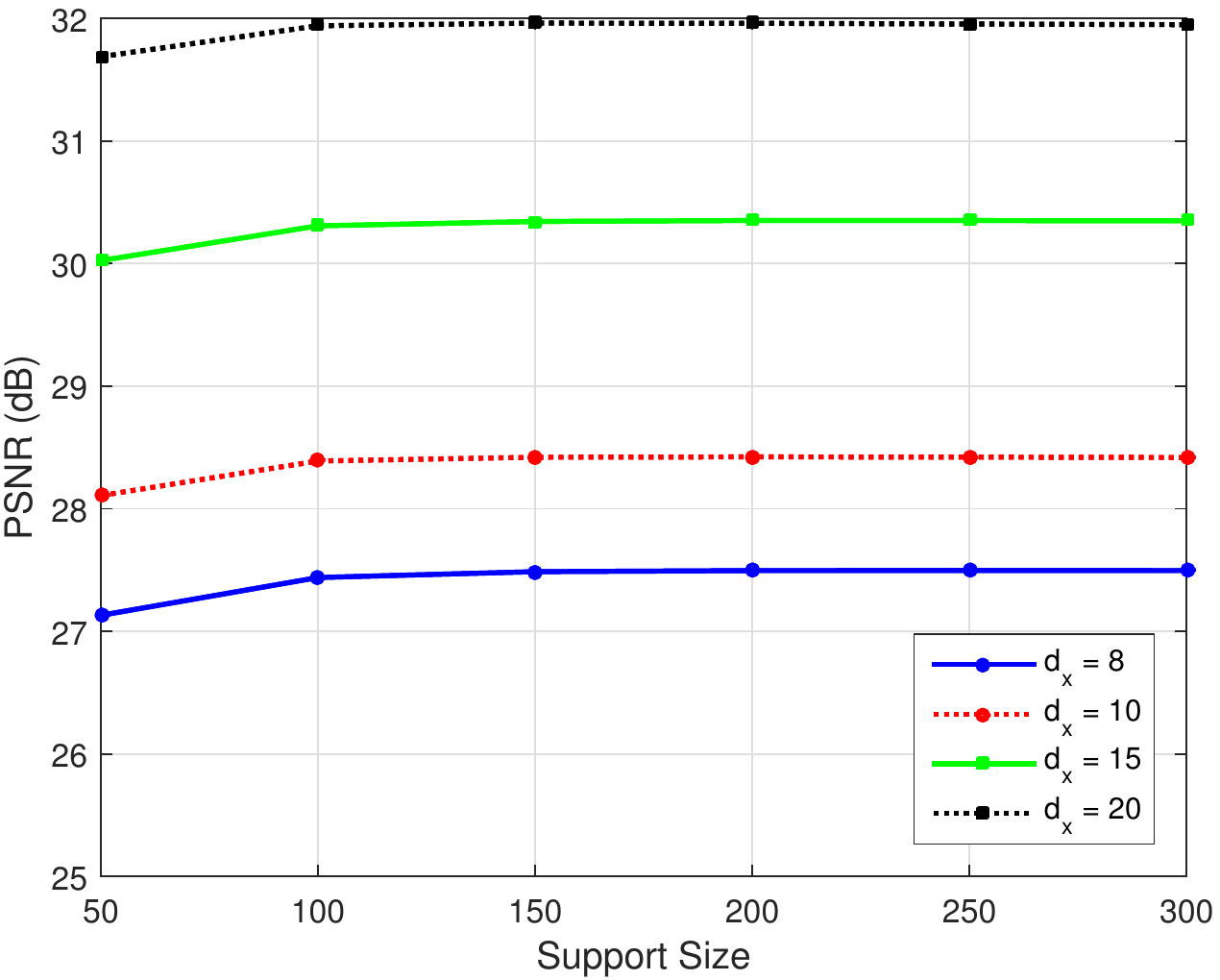}
  \caption{Quality Analysis}
  \label{fig:param_analysis_quality_analysis}
\end{subfigure}%
\begin{subfigure}{.5\textwidth}
  \centering
  \includegraphics[width=.95\linewidth]{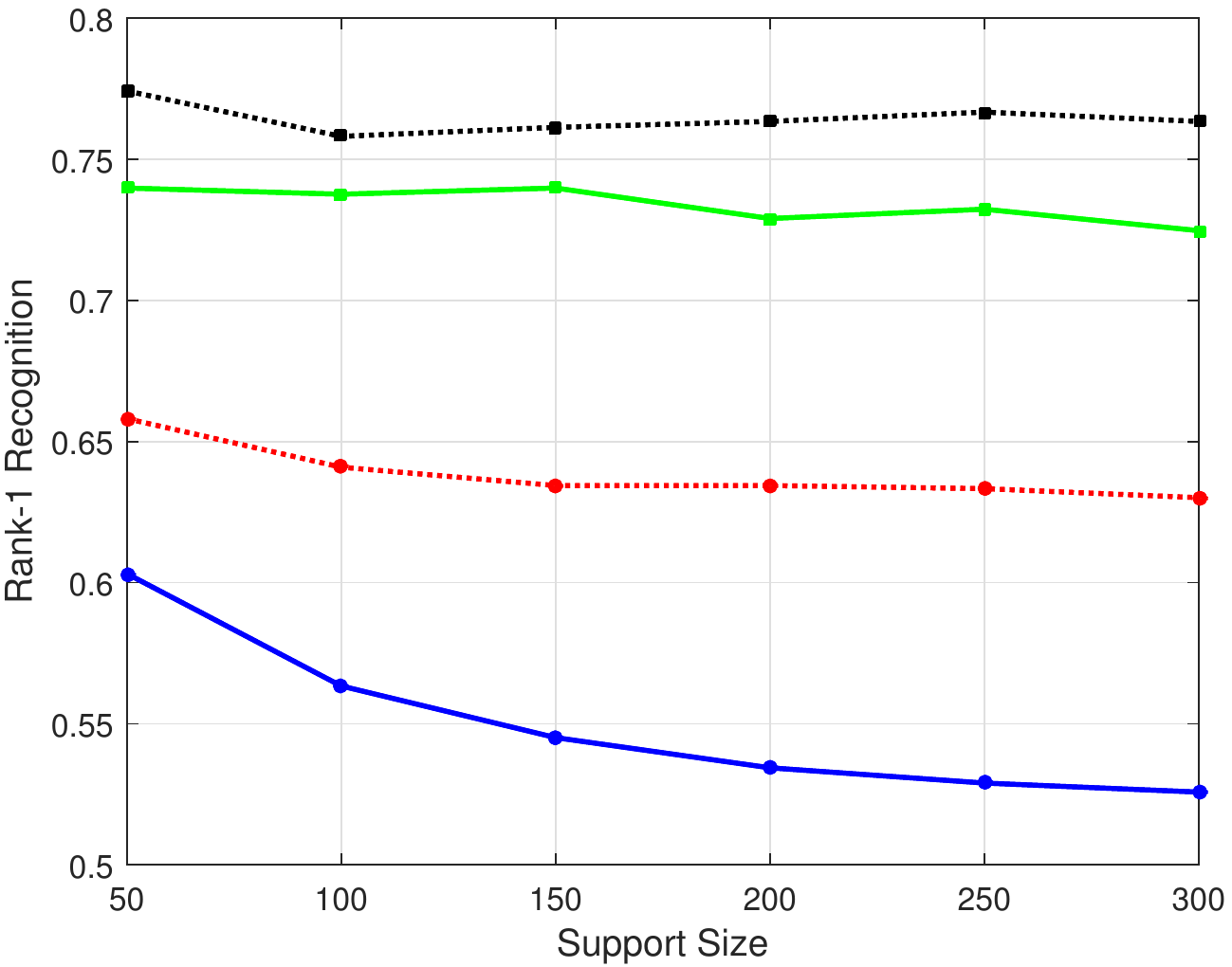}
  \caption{Recognition Analysis}
  \label{fig:param_analysis_rank1_analysis}
\end{subfigure}
\caption{Analysing the effect of the number of anchor vectors has on the performance of the proposed hallucination method.}
\label{fig:param_analysis}
\end{figure*}

In this section we investigate the effect of the number of anchor vectors $k$ of the proposed method. 
Fig. \ref{fig:param_analysis} shows the average PSNR and Rank-1 recognition using LBP face recognizer on all 930 probe images.
From these results it can be observed that PSNR increases as the number of neighbours is increased, until $k = 150$ where it starts decreasing (or stays in steady state). 
On the other hand, the best rank-1 recognition is attained at $k=50$, and recognition starts decreasing at larger values of $k$. 
This can be explained by the fact that increasing the number of support points will increase the number of high-resolution atoms to be combined (see \eqref{eq:phi_beta_equiv_2}), thus reducing the texture consistency of the hallucinated patch. More theoretical details are provided in section \ref{sec:proposed_method}.

\noindent

\subsection{Recognition Analysis}

The recognition performance of the proposed and other state-of-the-art face hallucination methods are summarized in Table \ref{tbl:table_recognition}.
In this table we adopt the Area Under the ROC curve (AUC) as a scalar-valued measure of accuracy for unsupervised learning \cite{huang_2015} together with the rank-1 recognition. It can be seen that the proposed method is most of the time superior to all the methods considered in this experiment when using the same averaging stitching method.
This performance can be further improved using the Quilting stitching method which provides better texture preservation properties.

\begin{table*}
\centering
\caption{Summary of the Rank-1 recognition results and Area Under Curve (AUC) metric using two different face recognition algorithms.}
\label{tbl:table_recognition}
\begin{tabular}{llcccccccc}
\thickhline
\multirow{3}{*}{\bf{Hall}$^\text{\bf{n}}$ \bf{Method}} & \multirow{3}{*}{\bf{Rec}$^\text{\bf{n}}$ \bf{Method}} & \multicolumn{8}{c}{\bf{Resolution $d_x$}} \\

                        &                         & \multicolumn{2}{c}{\bf{8}} & \multicolumn{2}{c}{\bf{10}} & \multicolumn{2}{c}{\bf{15}} & \multicolumn{2}{c}{\bf{20}}\\
                        &                         & \bf{rank-1} & \bf{AUC} & \bf{rank-1} & \bf{AUC} & \bf{rank-1} & \bf{AUC} & \bf{rank-1} & \bf{AUC} \\
\hline 
\multirow{2}{*}{Bi-Cubic} & Gabor & 0.0000 & 0.6985 & 0.0000 & 0.7823 & 0.0344 & 0.8829 & 0.5215 & 0.9181 \\
                          & LBP   & 0.3065 & 0.9380 & 0.5032 & 0.9598 & 0.6065 & 0.9708 & 0.7054 & 0.9792 \\
\hline  
\multirow{2}{*}{Eigentransformation \cite{wang_2005}} & Gabor & 0.0591 & 0.7852 & 0.1097 & 0.8359 & 0.3312 & 0.8841 & 0.5183 & 0.9098 \\
                                                     & LBP   & 0.2559 & 0.9390 & 0.4516 & 0.9554 & 0.5624 & 0.9633 & 0.6495 & 0.9688 \\
\hline                                                     
\multirow{2}{*}{Neighbour Embedding \cite{chang_2004}} & Gabor & 0.2323 & 0.8624 & 0.4710 & 0.8968 & 0.6172 & 0.9182 & 0.6409 & 0.9272  \\
                                                     & LBP   & 0.5548 & 0.9635 & 0.6398 & 0.9712 & 0.7215 & 0.9795 & 0.7559 & 0.9830 \\
\hline
\multirow{2}{*}{Sparse Position-Patches \cite{jung_2011}} & Gabor & 0.2333 & 0.8632 & 0.4645 & 0.8969 & 0.6118 & 0.9152 & 0.6398 & 0.9254  \\
& LBP   & 0.5677 & 0.9649 & 0.6441 & 0.9721 & 0.7247 & 0.9803 & 0.7570 & 0.9830 \\
\hline

\multirow{2}{*}{Position-Patches \cite{ma_2009}} & Gabor & 0.1108 & 0.8354 & 0.2849 & 0.8814 & 0.5774 & 0.9154 & 0.6419 & 0.9281  \\
                                                     & LBP   & 0.4699 & 0.9588 & 0.5849 & 0.9675 & 0.6849 & 0.9782 & 0.7312 & 0.9812 \\
\hline
\multirow{2}{*}{Eigen-Patches \cite{chen_2014}} & Gabor & 0.1613 & 0.8517 & 0.3849 & 0.8934 & 0.6065 & 0.9172 & 0.6387 & \bf{0.9283}  \\
                                                     & LBP   & 0.5226 & 0.9625 & 0.6215 & 0.9704 & 0.7237 & 0.9800 & 0.7602 & 0.9830 \\
\hline
\multirow{2}{*}{LINE \cite{jiang_2014}} & Gabor & 0.3118 & 0.8696 & 0.5011 & 0.8986 & 0.6118 & 0.9168 & 0.6409 & 0.9252  \\
                                                     & LBP   & 0.5925 & 0.9647 & 0.6559 & 0.9714 & 0.7323 & \bf{0.9804} & 0.7677 & \bf{0.9833} \\
\hline
\multirow{2}{*}{Proposed ($k = 50$)} & Gabor & 0.2753 & \bf{0.8803} & 0.5000 & 0.9036 & 0.6183 & 0.9202 & \bf{0.6452} & 0.9281  \\
                                                     & LBP   & 0.6032 & 0.9658 & 0.6581 & 0.9722 & \bf{0.7398} & 0.9798 & 0.7742 & \bf{0.9833} \\
\hdashline
\multirow{2}{*}{Proposed Quilt. ($k = 50$)} & Gabor & \bf{0.3258} & 0.8785 & \bf{0.5086} & \bf{0.9051} & \bf{0.6226} & \bf{0.9204} & 0.6409 &  0.9275 \\
                                     & LBP   & \bf{0.6065} & \bf{0.9663} & \bf{0.6656} & \bf{0.9732} & 0.7355 & 0.9799 & \bf{0.7753} & 0.9825 \\
\thickhline
\end{tabular}
\end{table*}

\begin{figure*}[htb]
\begin{subfigure}{.5\textwidth}
  \centering
  \includegraphics[width=.95\linewidth]{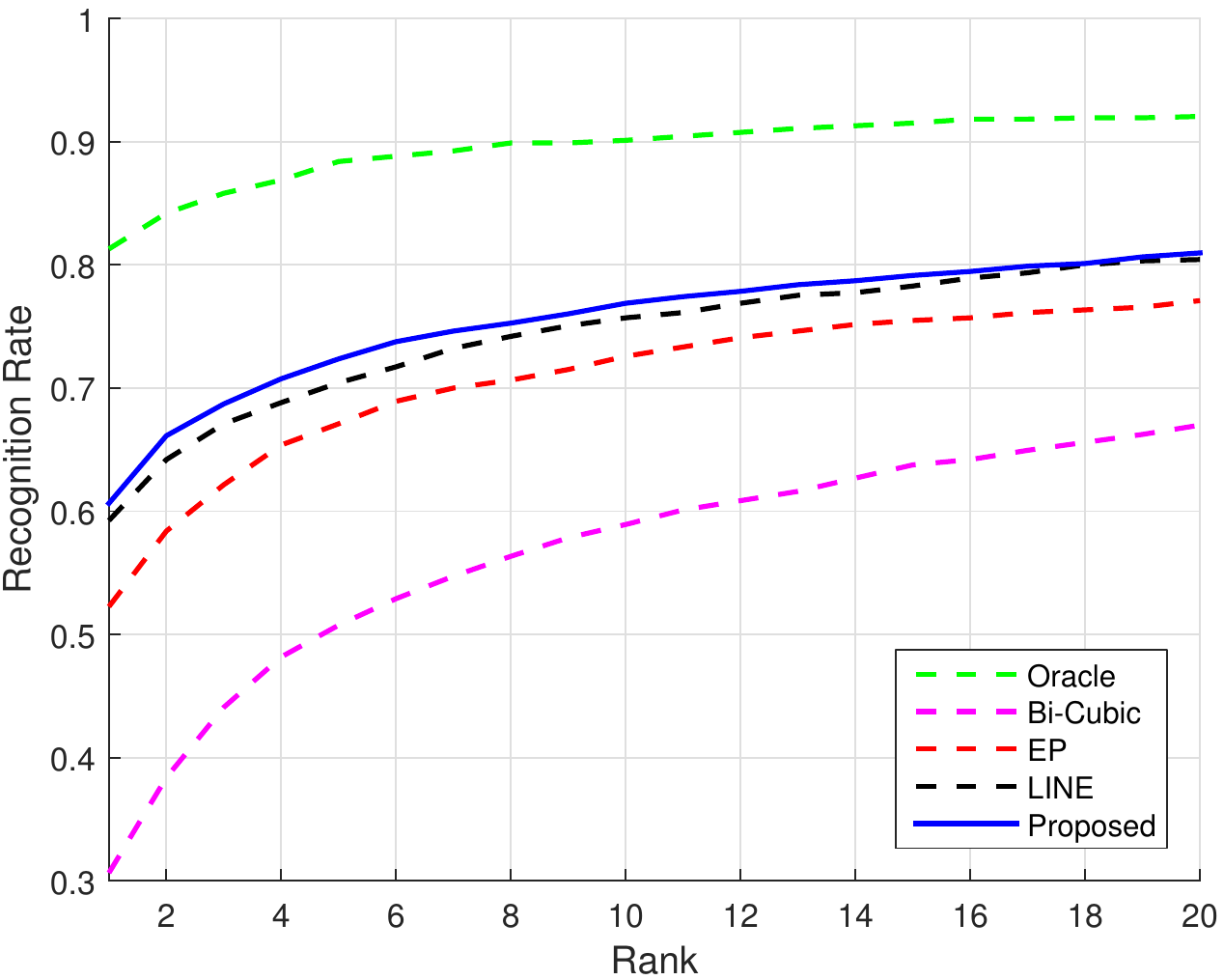}
  \caption{$d_x = 8$}
  \label{fig:cmc_dx_8}
\end{subfigure}%
\begin{subfigure}{.5\textwidth}
  \centering
  \includegraphics[width=.95\linewidth]{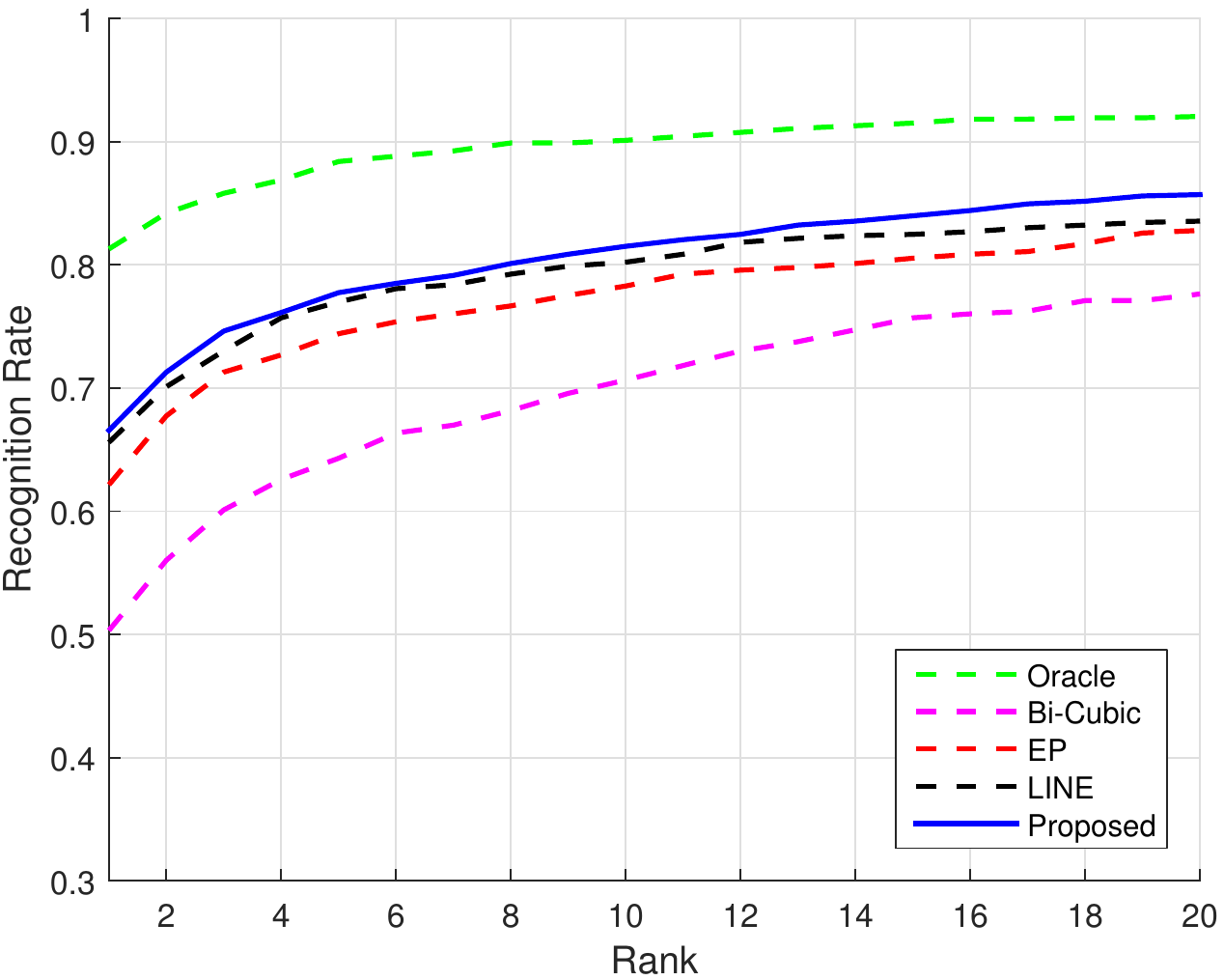}
  \caption{$d_x = 10$}
  \label{fig:cmc_dx_10}
\end{subfigure}%
\\
\begin{subfigure}{.5\textwidth}
  \centering
  \includegraphics[width=.95\linewidth]{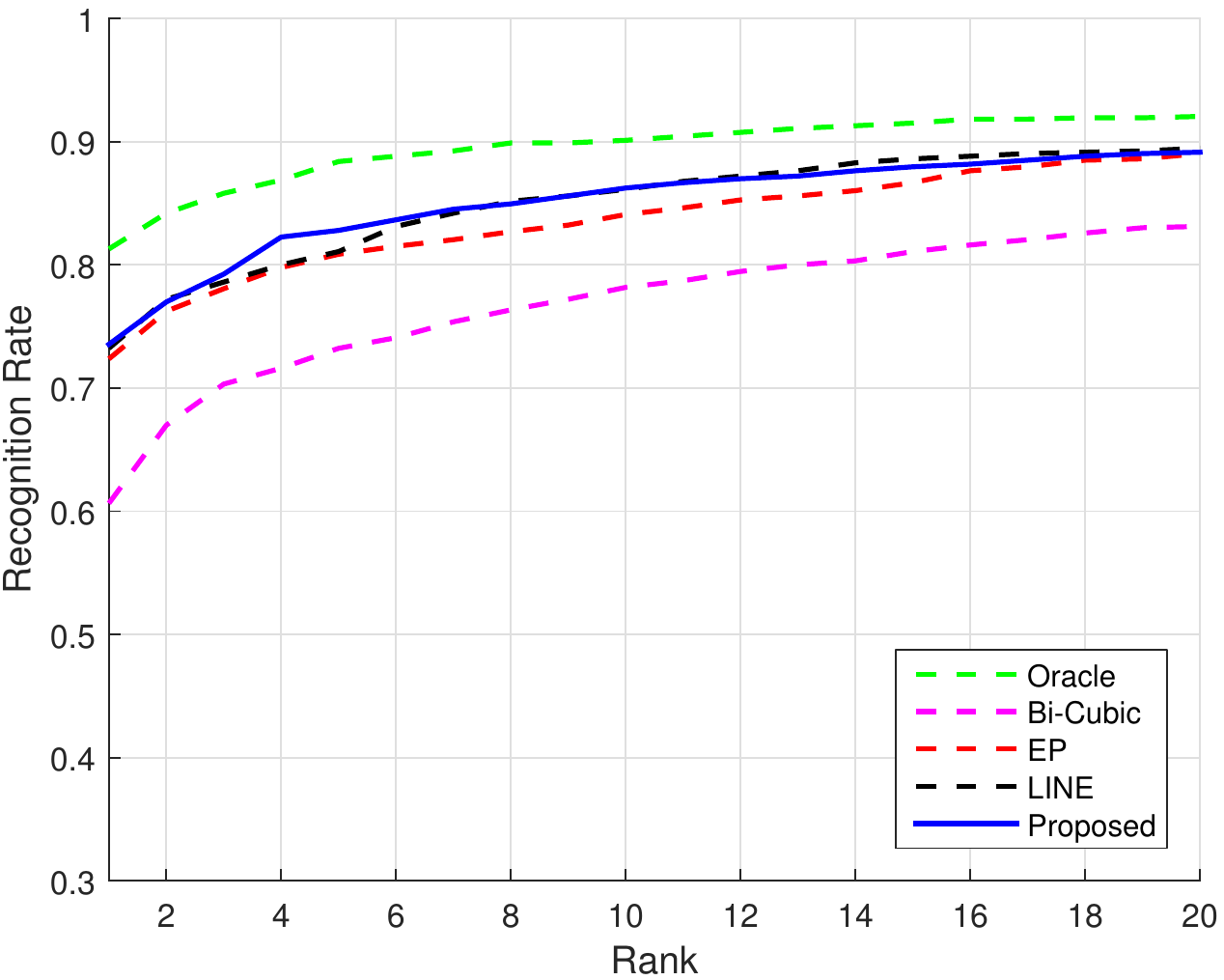}
  \caption{$d_x = 15$}
  \label{fig:cmc_dx_15}
\end{subfigure}
\begin{subfigure}{.5\textwidth}
  \centering
  \includegraphics[width=.95\linewidth]{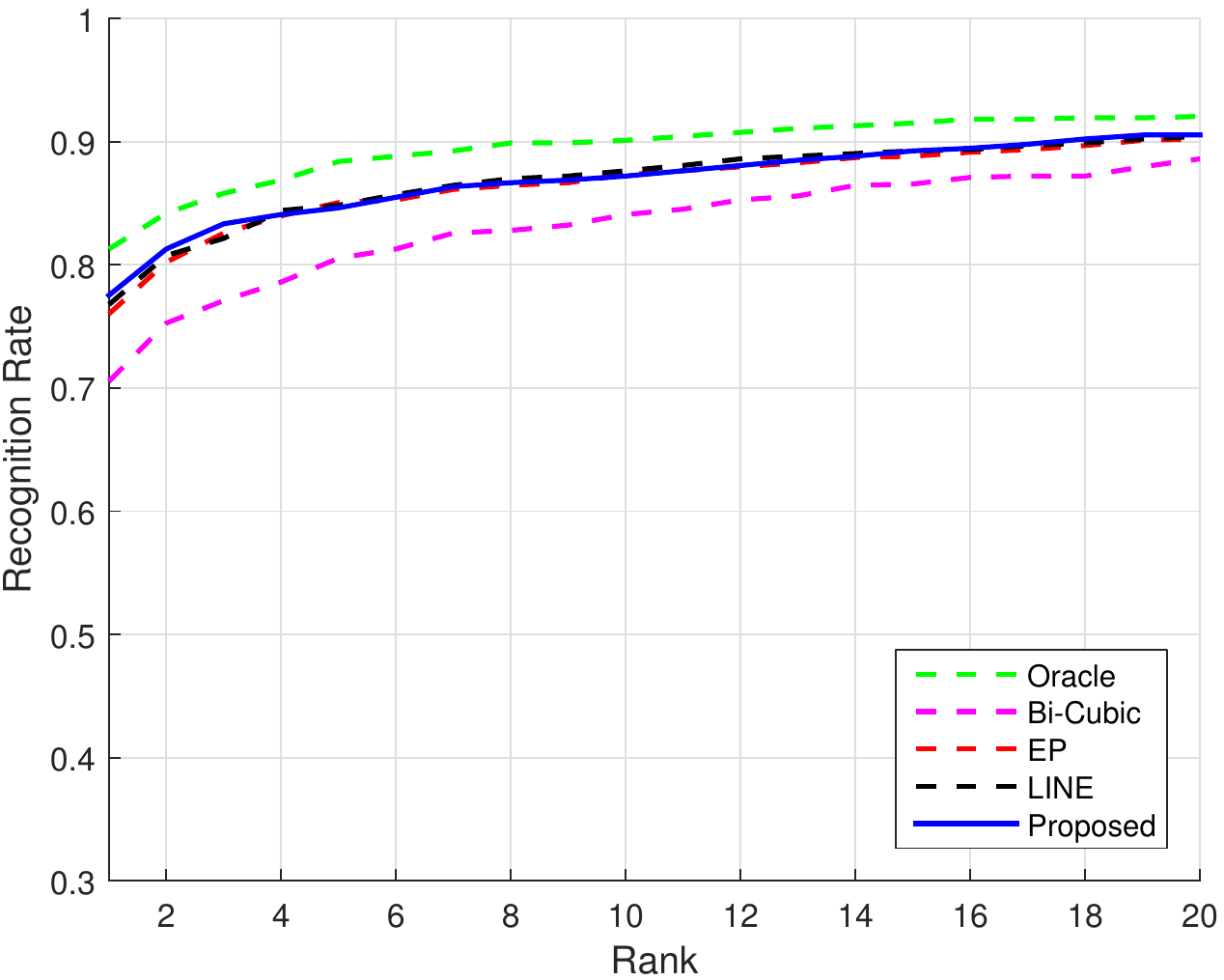}
  \caption{$d_x = 20$}
  \label{fig:cmc_dx_20}
\end{subfigure}
\caption{Cumulative Matching Score Curves (CMC) of face images hallucinated from different resolutions $d_x$.}
\label{fig:cmc_recognition_analysis}
\end{figure*}

Figure \ref{fig:cmc_recognition_analysis} shows the Cumulative Matching Score Curve (CMC) which measures the recognition at different ranks.
Because of the lack of space, only CMCs of LBP face recognizer were included.
For clarity, only LINE which provided the most competitive rank-1 recognition and Eigen-Patches, which as will be shown in the latter subsection, provides the most competitive results in terms of quality
were included, and compared to the oracle where the probe images were not down sampled.
Bi-cubic interpolation was included to show that state-of-the-art face recognition methods benefit from the texture details recovered by patch-based hallucination methods, achieving significant higher recognition rates at all ranks.
It can also be noted that the proposed method outperforms the other hallucination methods, especially at lower resolutions and lower ranks.

\noindent

\subsection{Quality Analysis}
\label{sec:quality_analysis}

Table \ref{tbl:table_quality} shows the quality analysis measured in terms of PSNR and SSIM.
These results reveal that the proposed method outperforms all other methods achieving PSNR gains of around 0.3 dB over LINE using the same neighbourhood size and 0.1 dB over Eigen-Patches which uses the whole dictionary, when using the same stitching method.
However, the use of quilting to stitch the patches provides a degradation in quality.
This can be explained since the averaging stitching acts as a denoising algorithm over overlapping regions which while reduce the texture detail (contributing to reducing the recognition performance), it suppresses the noise hallucinated in each patch.

\begin{table*}[!htbp]
\centering
\caption{Summary of the Quality Analysis results using the PSNR and SSIM quality metrics.}
\label{tbl:table_quality}
\begin{tabular}{lcccccccc}
\thickhline
\multirow{3}{*}{\bf{Hall}$^\text{\bf{n}}$ \bf{Method}} & \multicolumn{8}{c}{\bf{Resolution $d_x$}} \\
                        & \multicolumn{2}{c}{\bf{8}} & \multicolumn{2}{c}{\bf{10}} & \multicolumn{2}{c}{\bf{15}} & \multicolumn{2}{c}{\bf{20}}\\
                        & \bf{PSNR} & \bf{SSIM} & \bf{PSNR} & \bf{SSIM} & \bf{PSNR} & \bf{SSIM} & \bf{PSNR} & \bf{SSIM} \\
\hline
Bi-Cubic 							 & 24.0292 & 0.6224 & 26.2024 & 0.7338 & 25.2804 & 0.7094 & 28.6663 & 0.8531 \\
Eigentransformation \cite{wang_2005} & 24.3958 & 0.6496 & 26.8645 & 0.7504 & 24.9374 & 0.6724 & 27.7883 & 0.7892 \\
Neighbour Embedding \cite{chang_2004}& 26.9987 & 0.7533 & 27.9560 & 0.7973 & 29.9892 & 0.8714 & 31.6301 & 0.9122 \\
Position-Patches \cite{ma_2009}       & 27.3044 & 0.7731 & 28.2906 & 0.8145 & 30.1887 & 0.8785 & 31.7192 & 0.9143 \\
Sparse Position-Patches \cite{jung_2011}       & 27.2500 & 0.7666 & 28.2219 & 0.8100 & 30.1290 & 0.8767 & 31.7162 & 0.9146 \\
Eigen-Patches \cite{chen_2014}       & 27.3918 & 0.7778 & 28.3847 & \bf{0.8196} & 30.3118 & 0.8842 & 31.8986 & 0.9203 \\
LINE \cite{jiang_2014}               & 27.0927 & 0.7591 & 28.0253 & 0.8009 & 30.0471 & 0.8727 & 31.6970 & 0.9131 \\
Proposed ($k = 150$)                 & \bf{27.4866} & \bf{0.7802} & \bf{28.4200} & 0.8009 & \bf{30.3431} & \bf{0.8845} & \bf{31.9610} & \bf{0.9209} \\\hdashline
Proposed Quilt. ($k = 150$)                   & 27.3916 & 0.7762  & 28.3345 & 0.8178 & 30.2512 & 0.8815 & 31.8507 & 0.9185  \\
\thickhline
\end{tabular}
\end{table*}

\begin{figure*}[htb]
\begin{center}
\begin{tabular}{@{}ccccc@{}}
\includegraphics[height=25mm]{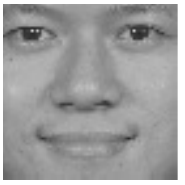} &
\includegraphics[height=25mm]{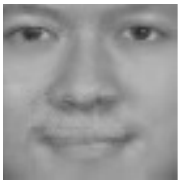} & 
\includegraphics[height=25mm]{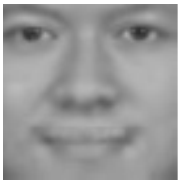} & 
\includegraphics[height=25mm]{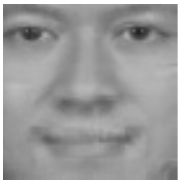} &
\includegraphics[height=25mm]{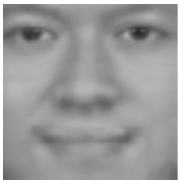} \\
\includegraphics[height=25mm]{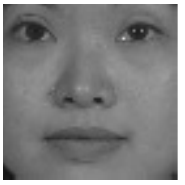} &
\includegraphics[height=25mm]{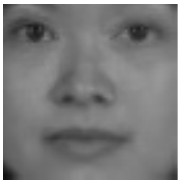} & 
\includegraphics[height=25mm]{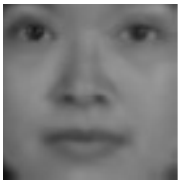} & 
\includegraphics[height=25mm]{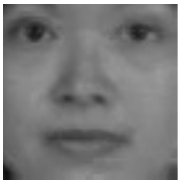} &
\includegraphics[height=25mm]{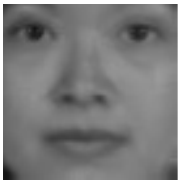} \\
\includegraphics[height=25mm]{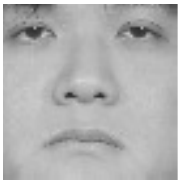} &
\includegraphics[height=25mm]{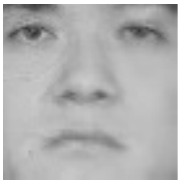} & 
\includegraphics[height=25mm]{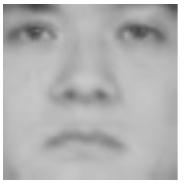} & 
\includegraphics[height=25mm]{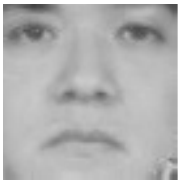} &
\includegraphics[height=25mm]{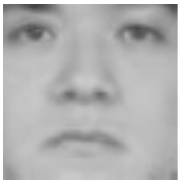} \\
\includegraphics[height=25mm]{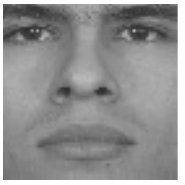} &
\includegraphics[height=25mm]{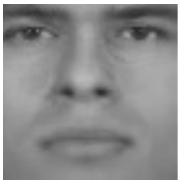} & 
\includegraphics[height=25mm]{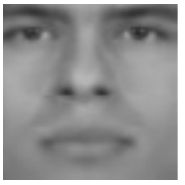} & 
\includegraphics[height=25mm]{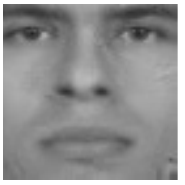} &
\includegraphics[height=25mm]{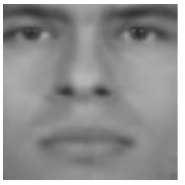} \\
\end{tabular}
\end{center}
\caption{Comparison of super-resolution results at resolution $d_x = 10$. From left to right are the high-resolution ground-truth, the results of LINE \cite{jiang_2014}, Eigen-Patches \cite{chen_2014}, the proposed method using $k=50$ and the proposed method with $k=150$.}
\label{fig:subjective_analysis_final}
\end{figure*}

Fig. \ref{fig:subjective_analysis_final} compares subjectively our method with LINE and Eigen-Patches, which were found to provide the most competitive results in terms of recognition and quality respectively.
It can be seen that Eigen-patches generally provides images which are blurred.
On the other hand, the images hallucinated by LINE contain more texture detail but have visible artefacts.
The images recovered by the proposed method manages to preserve more texture when having $k = 50$ at the expense of having more noise.
The results in table \ref{tbl:table_recognition} reveal that the face recognizers employed in this experiment are more robust to these distortions than to the smoothing effect provided by Eigen-Patches.
These results also reveal that these distortions can be alleviated by simply increasing the size of the support.

\subsection{Complexity Analysis}
\label{sec:complexity_analysis}

The complexity of the first layer which computes ridge regression using all $m$ entries in the dictionary is of order $O((1+\alpha^2)n^2(n+m))$. The second process first computed BPDN followed by Multivariate Ridge Regression on the selected $k$ anchor points. 
In this work we use the SparseLab solver for BPDN which employs Primal-Dual Interior-Point Algorithm whose complexity is of order $O(m^3)$ \cite{yang_2010}.
The complexity of Multivariate Ridge Regression using $k$ support vectors is of the order $O((1+\alpha^2)n^2(n+k))$.
This analysis reveals that the complexity of the proposed method is mostly dependent on the complexity of the sparse optimization method used.

\begin{table}[!htbp]
\centering
\caption{Summary of the time taken (in seconds) to synthesize one image at different resolutions.}
\label{tbl:table_complexity}
\begin{tabular}{lcccc}
\thickhline
\multirow{2}{*}{\bf{Hall}$^\text{\bf{n}}$ \bf{Method}} & \multicolumn{4}{c}{\bf{Resolution $d_x$}} \\
           & \bf{8} & \bf{10} & \bf{15} & \bf{20}\\\hline
Eigentransformation \cite{wang_2005}         & 2.84  & 2.69  & 2.74  & 2.75 \\
Neighbour Embedding \cite{chang_2004}        & 0.25  & 0.33  & 0.73  & 1.24 \\
Position-Patches \cite{ma_2009}              & 1.59  & 2.11  & 4.69  & 8.37 \\
Sparse Position-Patches \cite{jung_2011}     & 0.59  & 0.79  & 1.72  & 3.02 \\
Eigen-Patches \cite{chen_2014}               & 10.89 & 14.80 & 34.74 & 63.98 \\
LINE \cite{jiang_2014}                       & 2.23  & 2.16  & 2.61  & 2.83 \\
Proposed                                     & 8.04  & 5.78  & 4.71  & 5.43 \\
\thickhline
\end{tabular}
\end{table}

The complexity in terms of the average time taken to synthesize a high-resolution image from a low-resolution image in seconds is summarized in Table \ref{tbl:table_complexity}.
These results show that the proposed method is significantly less computationally intensive than Eigen-Patches but more complex than the other methods, including LINE.
While complexity is not the prime aim of this work, the performance of the proposed scheme can be significantly improved using more efficient $l_1$-minimization algorithms as mentioned in \cite{yang_2010}.

\section{Conclusion}
\label{sec:conclusion}

In this paper, we propose a new approach which can be used to super-resolve a high-resolution image from a low-resolution test image.
The proposed method first derives a smooth approximation of the ground-truth on the high-resolution manifold using all entries in the coupled dictionaries.
Then we use Basis Pursuit Denoising to find the optimal atomic decomposition to represent the approximated sample $\widetilde{\mathbf{y}}_i^{s\{0\}}$.
Based on the assumption that the patches reside on a high-resolution manifold, we assume that the optimal support to represent $\widetilde{\mathbf{y}}_i^{s\{0\}}$ is good to reconstruct the ground-truth, and use the coupled support to hallucinate the high-resolution patch using Multivariate Ridge Regression.
Extensive simulation results demonstrate that the proposed method outperforms the six face super-resolution methods in both recognition and quality analysis.

Future work points us in the direction to implement face hallucination techniques which are able to hallucinate and enhance face images afflicted by different distortions such as compression, landmark-point misalignment, bad exposure and other distortions commonly found in CCTV images.
The ability of current schemes (including the proposed method) are dependent on the dictionaries used, and therefore these schemes can be made more robust by building more robust dictionaries.


%

\appendices
\section{Indirect and Direct Projection Matrices}
\label{sec:proof_1}

Using the results from section \ref{sec:mvrr}, the direct up-scaling projection matrix is defined by

\begin{equation}
\mathbf{\Phi}_i = \mathbf{H}_i \mathbf{L}_i^T \left( \mathbf{L}_i\mathbf{L}_i^T + \lambda \mathbf{I}\right)^{-1}
\label{eq:appendix_direct}
\end{equation}

\noindent while the indirect projection matrix according to the authors in \cite{zhang_2015} is given by

\begin{equation}
\mathbf{\Phi}_i^\zeta = \mathbf{H}_i \left( \mathbf{L}_i^T \mathbf{L}_i + \lambda \mathbf{I}\right)^{-1}\mathbf{L}_i^T
\label{eq:appendix_indirect}
\end{equation}

The covariance matrix $\mathbf{C} =  \mathbf{L}_i\mathbf{L}_i^T \in \mathbb{R}^{n \times n}$ while the covariance matrix $\mathbf{C}^\zeta = \mathbf{L}_i^T \mathbf{L}_i \in \mathbb{R}^{m \times m}$, where $\rank{(\mathbf{C})} = \rank{(\mathbf{C}^\zeta)} \le n$. This shows that $\mathbf{C}^\zeta$ has at least $m-n$ zero eigen-values.
Given the above observation, we hypothesize that if the solutions in \eqref{eq:appendix_direct} and \eqref{eq:appendix_indirect} are equivalent, then

\begin{equation}
  \begin{aligned}
        \left( \mathbf{L}_i^T \mathbf{L}_i + \lambda \mathbf{I}\right)^{-1}\mathbf{L}_i^T &= \mathbf{L}_i^T \left( \mathbf{L}_i\mathbf{L}_i^T + \lambda \mathbf{I}\right)^{-1} \\
        \mathbf{L}_i^T \left( \mathbf{L}_i\mathbf{L}_i^T + \lambda \mathbf{I}\right) &= \left( \mathbf{L}_i^T \mathbf{L}_i + \lambda \mathbf{I}\right) \mathbf{L}_i^T \\
\mathbf{L}_i^T \mathbf{L}_i\mathbf{L}_i^T + \lambda \mathbf{L}_i^T \mathbf{I} &= \mathbf{L}_i^T \mathbf{L}_i \mathbf{L}_i^T + \lambda \mathbf{L}_i^T \mathbf{I}
  \end{aligned}  		
\end{equation}

\noindent Therefore, we can conclude that the direct and indirect projection matrices are equivalent.


\section*{Acknowledgment}

The authors would like to thank Junjun Jiang who is the author in \cite{jiang_2014} and Vitomir \v{S}truc author in \cite{struc_2010} for proving the source code of their methods.
The authors would also like to thank Prof. David Donoho and his team at SparseLab for providing the BPDN $l_1$-minimization solver.

\ifCLASSOPTIONcaptionsoff
  \newpage
\fi



%



\bibliographystyle{IEEEtran}
\bibliography{IEEEabrv,face_SR}
%

\begin{IEEEbiography}[{\includegraphics[width=1in,height=1.25in,clip,keepaspectratio]{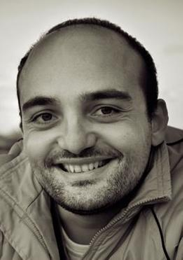}}]{Reuben A. Farrugia}
(S’04, M’09) received the first
degree in Electrical Engineering from the University of
Malta, Malta, in 2004, and the Ph.D. degree from the
University of Malta, Malta, in 2009.
In January 2008 he was appointed Assistant Lecturer with the same department
and is now a Senior Lecturer. 
He has been in technical and organizational committees of several national and international conferences. In particular, he served. as General-Chair on the IEEE Int. Workshop on Biometrics and Forensics (IWBF) and as Technical Programme Co-Chair on the IEEE Visual Communications and Image Processing (VCIP) in 2014.  He has been contriuting as a reviewer of several journals and counferences, including IEEE Transactions on Image Processing, IEEE Transactions on Circuits and Systems for Video and Technology and IEEE Transactions on Multimedia. On September 2013 he was appointed as National Contact Point of the European Association of Biometrics (EAB).
\end{IEEEbiography}

\begin{IEEEbiography}[{\includegraphics[width=1in,height=1.25in,clip,keepaspectratio]{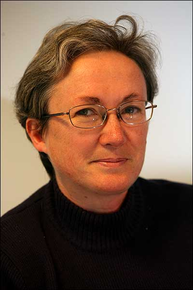}}] {Christine Guillemot} IEEE fellow, is “Director of Research” at INRIA, head of a research team
dealing with image and video modeling, processing, coding and communication. She holds a Ph.D. degree from ENST
(Ecole Nationale Superieure des Telecommunications) Paris, and an “Habilitation for Research Direction” from the
University of Rennes. From 1985 to Oct. 1997, she has been with FRANCE TELECOM, where she has been involved
in various projects in the area of image and video coding for TV, HDTV and multimedia. From Jan. 1990 to mid
1991, she has worked at Bellcore, NJ, USA, as a visiting scientist. She has (co)-authored 24 patents, 9 book chapters,
60 journal papers and 140 conference papers. She has served as associated editor (AE) for the IEEE Trans. on Image
processing (2000-2003), for IEEE Trans. on Circuits and Systems for Video Technology (2004-2006) and for IEEE Trans. On Signal Processing
(2007-2009). She is currently AE for the Eurasip journal on image communication, IEEE Trans. on Image Processing (2014-2016) and member of the editorial board for the IEEE Journal on
selected topics in signal processing (2013-2015). She is a member of the IEEE IVMSP technical committee.
\end{IEEEbiography}




\end{document}